\pdfoutput=1

\documentclass[11pt]{article}

\usepackage[preprint]{acl}

\usepackage{times}
\usepackage{latexsym}

\usepackage[T1]{fontenc}

\usepackage[utf8]{inputenc}

\usepackage{microtype}

\usepackage{inconsolata}

\usepackage{graphicx}
\usepackage{booktabs}
\usepackage{multirow}
\usepackage{amsmath}
\usepackage{xurl}

\title{The State of Commercial Automatic French Legal Speech \\Recognition Systems and their Impact on Court Reporters et al.}

\author{Nicolas Garneau \\
  University of Copenhagen \\
  \texttt{nicolas@nlp.quebec} \\\And
  Olivier Bolduc \\
  P.O.B. Sténographes officiels \\
  \texttt{obolduc@stenopob.ca} \\}

\begin{document}
\maketitle
\begin{abstract}
In Quebec and Canadian courts, the transcription of court proceedings is a critical task for appeal purposes and must be certified by an official court reporter. The limited availability of qualified reporters and the high costs associated with manual transcription underscore the need for more efficient solutions. This paper examines the potential of Automatic Speech Recognition (ASR) systems to assist court reporters in transcribing legal proceedings. We benchmark three ASR models, including commercial and open-source options, on their ability to recognize French legal speech using a curated dataset. Our study evaluates the performance of these systems using the Word Error Rate (WER) metric and introduces the Sonnex Distance to account for phonetic accuracy. We also explore the broader implications of ASR adoption on court reporters, copyists, the legal system, and litigants, identifying both positive and negative impacts. The findings suggest that while current ASR systems show promise, they require further refinement to meet the specific needs of the legal domain.
\end{abstract}



\section{Introduction}

In Quebec and Canadian courts, as in most North American jurisdictions, court proceedings may be transcribed, particularly for appeal purposes.
The transcription can either be done live if a court reporter is present, or offline if the proceedings have been recorded.
Needless to say, a limited number of qualified people are available to carry out this type of work in person, hence most of the transcriptions are done offline using recordings.
In both cases, to be admissible in evidence, these transcripts must be certified by an official court reporter, giving them a quasi-authentic character.
As late as 2024, the task of transcribing offline court hearings is carried out by copyists working under the supervision of an official court reporter, from audio files prepared by courthouse administrative staff. These recordings are either hand-delivered or sent by mail, on CD-ROM, to be typed out entirely by hand.
The costs associated with these transcriptions represent a significant expense for the Department of Justice, as they are required by the agencies in its portfolio as well as by the courts and legal affairs.
Given the recent advent in natural language processing and large language models, it is natural to wonder what is the impact of such technologies on the profession of court reporters and copyists, as well as the courthouses and the litigants.
In this paper, we study how Automatic Speech Recognition (ASR) systems available on the market can help in producing transcripts of court hearings.
The task of ASR has long been studied, hence the availability of different pre-trained models on the market~\cite{hannun2014deep, chan2016listen, amodei2016deep, baevski2020wav2vec}.
There are commercial and open-source ASR models and in this paper, similar to what \citet{miner2020assessing} did in the field of psychotherapy, we wish to benchmark three different models on the automatic French legal speech recognition task using a curated benchmark.

\section{Automatic Speech Recognition Models}

Each model considered in this study offers a French Canadian configuration to perform ASR from a source audio.
We suppose that all of these models rely on state-of-the-art deep neural transformer networks~\cite{Vaswani2017AttentionIA, baevski2020wav2vec, chan2021speechstew}.

\paragraph{Amazon Web Services.} Amazon Web Services (AWS) offers Amazon Transcribe as an ASR tool.
The user needs to upload its audio file in the console and issue a transcription job to receive the text file.
At the time of this writing, it costs 0.024 USD per minute for the first 250,000 minutes. AWS offers only one model for the French Canadian language\footnote{https://aws.amazon.com/fr/transcribe/}.

\paragraph{Google Cloud Platform.} The Google Cloud Platform (GCP) offers a Speech-to-Text tool.
Similar to AWS, the user needs to upload its audio file and issue a transcription job to receive the text file.
In this study, we consider the second version of their API.
At the time of this writing, it costs 0.016 USD per minute for the first 500,000 minutes.
GCP offers different multilingual models, and we selected the latest Chirp 2 in our benchmark\footnote{\url{https://cloud.google.com/speech-to-text/?hl=en}}~\cite{zhang2023google}.


\paragraph{OpenAI Whisper.} OpenAI offers Whisper (OW), an open-source ASR model available through HuggingFace which has been initially published by \citet{radford2023robust}.
In this study, we consider the medium multilingual version of Whisper which can easily fit on a GPU and process long audio files\footnote{\url{https://huggingface.co/openai/whisper-medium}}.
One of the main advantages of Whisper is that the user has full control over the model, and can use it on its computer.
However, this can be a disadvantage if the user is not acquainted with such technologies, and a graphical interface can become more handy.
Moreover, the need for a GPU can be costly for a self-employed stenographer.
It is worth noting that Microsoft Word, a software widely used by stenographers and copyists, offers built-in automatic speech recognition.
While they do not disclose the exact model used for the transcription, we suspect that they are using OpenAI Whisper Large V2 since this model is proposed by their cloud solution~\footnote{\url{https://azure.microsoft.com/en-us/blog/accelerate-your-productivity-with-the-whisper-model-in-azure-ai-now-generally-available/}}.

\section{Benchmark Dataset}

The benchmark dataset is composed of two types of audio files; the first one is the Commission Charbonneau which happened from 2012 until 2015\footnote{\url{https://www.bibliotheque.assnat.qc.ca/guides/fr/les-commissions-d-enquete-au-quebec-depuis-1867/7732-commission-charbonneau-2015}}.
These recordings are mainly testimonies and interviews from different people involved in the Commission.
This dataset includes 1 day, 0 hours, 22 minutes, and 2.05 seconds of recordings.
The second type is judgments read out loud by a judge.
There may be interactions between the lawyers, court clerks, and the litigants.
This dataset includes 16 judgments totaling 6 hours, 15 minutes, and 55.55 seconds of recording.
In total, we have 1 day, 6 hours, 37 minutes, and 57.6 seconds of long-form audio to be transcribed by the systems under test.
For each of these recordings, we have access to professional transcriptions made by court reporters.

\section{Experiments}

\begin{table*}[h!]
    \centering
    \resizebox{\textwidth}{!}{
    \begin{tabular}{l|ccccc|ccccc|ccccc}
        \multicolumn{1}{c}{} & \multicolumn{5}{c}{\textbf{Judgements}} & \multicolumn{5}{c}{\textbf{Commission Charbonneau}} & \multicolumn{5}{c}{\textbf{Both}} \\
       \textbf{Model}  & \textbf{Ins.} &  \textbf{Del.} & \textbf{Subst.} & \textbf{WER} & \textbf{\#Words} & \textbf{Ins.} &  \textbf{Del.} & \textbf{Subst.} & \textbf{WER} & \textbf{\#Words} & \textbf{Ins.} &  \textbf{Del.} & \textbf{Subst.} & \textbf{WER} & \textbf{\#Words}\\
       \midrule
       AWS  & 102 & \textbf{149} & \underline{329} & 0.15 & \multirow{3}*{3823} & 302 & \textbf{369} & \textbf{461} & \textbf{0.14} & \multirow{3}*{8143} & 231 & \textbf{291} & \textbf{414} & \textbf{0.15} & \multirow{3}*{6607}\\
       GCP & \underline{129} & \underline{242} & 305 & \underline{0.18} & & \underline{411} & 516 & 536 & 0.18 & & \underline{311} & \underline{419} & 453 & \underline{0.18} & \\
       OW & \textbf{39} & 171 & \textbf{238} & \textbf{0.12} & & \textbf{239} & \underline{530} & \underline{644} & 0.18 & & \textbf{168} & 403 & \underline{499} & 0.16 &  \\
    \end{tabular}
    }
    \caption{Number of insertions, deletions, substitutions, as well as word error rate per model on the Judgments and the Commission Charbonneau corpus. We provide the average number of words per document, as well as the statistics for the combination of both.}
    \label{tab:wer}
\end{table*}

\begin{figure}
    \centering
    \includegraphics[width=0.99\linewidth]{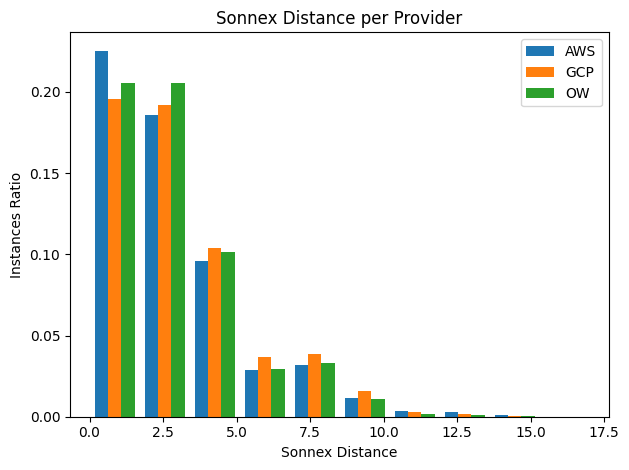}
    \caption{Ratio of instances and their Sonnex distance on the substitutions for each provider.}
    \label{fig:sonnex}
\end{figure}

The evaluation consists of comparing the generated transcripts by the different ASR systems to the one written by a court reporter.
We perform automatic evaluation using the widely adopted Word Error Rate (WER) metric to assess the performance of the different ASR systems.
The Word Error Rate is computed as follows;
\begin{equation}
    \text{WER} = \frac{I + D + S}{N}
\end{equation}

where $I$ is the number of insertions, $D$ the number of deletions, $S$ the number of substitutions, and $N$ the number of words in the official transcript.
Results of the three systems under test on the two corpora are displayed in Table~\ref{tab:wer}.
AWS provides overall the best performance, with an average WER of 0.15 on both corpora.
It is the system producing much fewer deletions and substitutions than its peers on the Commission Charbonneau corpus.
On the other hand, OW shines on the number of insertions, making very few of them.
It also has a surprisingly low WER on the judgment corpus.
The WER metric does not take into account homophones and treats substitutions like \textit{présente} $\rightarrow$ \textit{présentes} (plural) and \textit{présente} $\rightarrow$ \textit{avocat} equally even though the first pair sound the same whereas the second one sounds completely different.
To this end, we introduce the Sonnex Distance on the substitutions by obtaining the French phonemes of each word and computing the Levhenstein distance on the phonemes.
The results are displayed in Figure~\ref{fig:sonnex}.
AWS performs best with an average distance of X, followed by GCP and OW with an average distance of X and X respectively.
After closely looking at the results in the first bin (Sonnex distance ($< 2.5$), we found out that the types of errors effectively revolve around gender and number as well as verb tense and homophones. In other bins ($> 2.5$), we mostly see Proper noun errors as well as slang and dialects.
In total, it costs about 45 US dollars to run the experiment with AWS and 30 US dollars with GCP.
With OpenAI Whisper, we used a n1-standard-8 instance with a Tesla V100 GPU on the Google Cloud Platform for around 8 hours, which cost 23 US dollars to conduct all the experiments.
While using OpenAI Whisper seems to be the cheapest, it does require engineering skills to properly setup the virtual machine.
All things considered, every tested model is a great contender to produce pre-generated transcripts.
They all still have a fair amount of errors ($>15\%$), hence the need to carefully review these generations.





\section{Impacts}

We identify four actors that might be impacted differently with the hypothesis that ASR systems are approaching near-perfect transcriptions.
The first actor is inevitably the court reporter, who must approve and certify the transcriptions' correctness.
The second actor is the copyist, often employed by the court reporter to generate a pre-transcription of the audio.
The third actor is the legal system, which employs court reporters to produce the transcripts.
The last factor is the litigant, facing the legal system when involved in a legal dispute.

\subsection{Positive Impacts}
We identify several positive impacts of ASR systems on these four actors.

\paragraph{Efficiency.}
ASR systems can transcribe court proceedings in real time, and court reporters can focus on verifying and editing ASR-generated transcripts rather than typing everything manually or hiring a copyist, improving overall efficiency.
Assuming that ASR-generated transcripts really improves the efficiency of the court reporters overall, this will allow for faster delivery of final documents, clear existing backlogs, speeding up the legal process and reducing case delays.

\paragraph{Accuracy.} ASR systems can potentially difficult audio conditions and multiple speakers better over time, improving transcription quality. They can also reduce the likelihood of errors, leading to more accurate records. From a consistency perspective, ASR systems may provide uniform standard for transcriptions, improving the quality and reliability of legal records.

\paragraph{Cost Savings.} Courts may save money on transcription services as ASR can reduce the need for human labor. However, it is important to note that these savings can be redirected toward technology maintenance, training, and other areas.
Moreover, like many private players in the justice system, stenographers are remunerated on a unitary basis. The deployment of ASR-type solutions should therefore generate cost savings for transcribing subcontractors. In the current legal and regulatory context, these savings would be captured by the stenographers themselves, resulting in productivity gains through the reallocation of time/money resources.

\paragraph{Access to Justice.} Faster availability of transcripts can help expedite hearings and appeals, providing timely access to justice. Moreover, affordable transcription services can make legal processes more accessible to individuals and small businesses. This will also foster digital access since there are lots of audio that could not be transcribed due to a lack of manpower.

\paragraph{Fairness and Transparency.} Faster and more affordable transcription services can help level the playing field, giving all litigants, regardless of financial resources, equal access to accurate legal records. It can also enhance transparency in legal proceedings, ensuring that all parties have access to the same information.

\subsection{Negative Impacts}
However, it may come with several negative impacts that we need to comprehend before using these ASR systems in the legal system.
The specialized knowledge required to produce accurate transcriptions from a complex and diverse language will likely become scarce. Gains in transcription accuracy could plateau much more quickly than gains in productivity. Once productivity gains have been realized and maximized, a possible shortage of manpower to validate ASR model generations would result in a loss of transcript quality.

\paragraph{Job Displacement.} The biggest loser here is the copyist, who may need to retrain or acquire new skills to remain valuable in the evolving job market. Indeed, the need for a copyist may become less relevant as ASR systems handle the initial drafting. Court reporters may need to adapt to new tasks, such as editing ASR-generated transcripts. The profession may shift focus from transcription to quality control and legal knowledge.

\paragraph{Dependency on Technology.} Over-reliance on ASR systems could lead to issues if the technology fails or produces errors in complex cases. Courts will need contingency plans to handle transcription needs if the ASR system encounters problems or is unavailable.

\paragraph{Privacy and Security.} ASR systems handling sensitive legal information must be secure to prevent data breaches.
Thus, ensuring the confidentiality of court records becomes paramount, requiring robust cybersecurity measures, whether the ASR systems are on-premise or in the cloud. Litigants may have concerns about the privacy and security of their data when handled by ASR systems, hence maintaining transparency on how these systems handle and protect data will foster trust.

\paragraph{Costs and Maintenance.} It is true that ASR systems might cut costs on some of the manpower required to perform the transcriptions, but these systems need to be maintained by skilled engineers, which can also be costly. Indeed, ongoing maintenance and regular updates are necessary to ensure ASR systems remain accurate and reliable.

In light of these pros and cons, the jobs of copyists are definitely at stake.
Court reporters might need to shift their usual tasks as editors of pre-generated transcripts.
The legal systems will most likely benefit from ASR systems with cost savings and efficiency.
ASR systems will also foster access to justice for litigants by making this service cheaper to acquire.

\section{Conclusion}

The integration of ASR systems in the legal domain presents a significant opportunity to enhance the efficiency and accuracy of court transcription services.
Our benchmark study of three ASR models demonstrates the potential of these technologies to assist court reporters by providing pre-generated transcriptions.
However, the current models still face challenges in achieving the high accuracy required for legal documentation, particularly in handling complex audio conditions and multiple speakers.
While ASR systems can reduce the workload of court reporters and lower the costs associated with manual transcription, they also introduce new challenges.
The potential for job displacement among copyists and the need for court reporters to adapt to new roles focused on editing and quality control are significant considerations.
Additionally, issues related to privacy, data security, and the reliability of ASR systems in critical legal contexts must be addressed to ensure trust and adoption.
To fully realize the benefits of ASR in the legal field, further specialization and refinement of these models are necessary.
Developing ASR systems tailored to the nuances of legal speech and improving their accuracy will support court reporters in their essential role, thereby enhancing the transparency and efficiency of the judicial process.
The legal system stands to benefit from cost savings, while litigants will experience more timely and accurate access to legal records.
Ultimately, the successful integration of ASR technology in court reporting hinges on a balanced approach that leverages technological advancements while addressing the associated human and ethical impacts.
Finally, transcripts are official documents whose accuracy must be ensured to prevent miscarriages of justice. People's rights depend on it. In the short term, the immediate savings generated by implementing the ASR model should be reinvested, at least in part, in training and professional supervision to protect the integrity of the justice system and public confidence.

\bibliography{acl_latex}

\begin{thebibliography}{9}
\providecommand{\natexlab}[1]{#1}

\bibitem[{Amodei et~al.(2016)Amodei, Ananthanarayanan, Anubhai, Bai, Battenberg, Case, Casper, Catanzaro, Cheng, Chen et~al.}]{amodei2016deep}
Dario Amodei, Sundaram Ananthanarayanan, Rishita Anubhai, Jingliang Bai, Eric Battenberg, Carl Case, Jared Casper, Bryan Catanzaro, Qiang Cheng, Guoliang Chen, et~al. 2016.
\newblock Deep speech 2: End-to-end speech recognition in english and mandarin.
\newblock In \emph{International conference on machine learning}, pages 173--182. PMLR.

\bibitem[{Baevski et~al.(2020)Baevski, Zhou, Mohamed, and Auli}]{baevski2020wav2vec}
Alexei Baevski, Yuhao Zhou, Abdelrahman Mohamed, and Michael Auli. 2020.
\newblock wav2vec 2.0: A framework for self-supervised learning of speech representations.
\newblock \emph{Advances in neural information processing systems}, 33:12449--12460.

\bibitem[{Chan et~al.(2016)Chan, Jaitly, Le, and Vinyals}]{chan2016listen}
William Chan, Navdeep Jaitly, Quoc Le, and Oriol Vinyals. 2016.
\newblock Listen, attend and spell: A neural network for large vocabulary conversational speech recognition.
\newblock In \emph{2016 IEEE international conference on acoustics, speech and signal processing (ICASSP)}, pages 4960--4964. IEEE.

\bibitem[{Chan et~al.(2021)Chan, Park, Lee, Zhang, Le, and Norouzi}]{chan2021speechstew}
William Chan, Daniel Park, Chris Lee, Yu~Zhang, Quoc Le, and Mohammad Norouzi. 2021.
\newblock Speechstew: Simply mix all available speech recognition data to train one large neural network.
\newblock \emph{arXiv preprint arXiv:2104.02133}.

\bibitem[{Hannun et~al.(2014)Hannun, Case, Casper, Catanzaro, Diamos, Elsen, Prenger, Satheesh, Sengupta, Coates et~al.}]{hannun2014deep}
Awni Hannun, Carl Case, Jared Casper, Bryan Catanzaro, Greg Diamos, Erich Elsen, Ryan Prenger, Sanjeev Satheesh, Shubho Sengupta, Adam Coates, et~al. 2014.
\newblock Deep speech: Scaling up end-to-end speech recognition.
\newblock \emph{arXiv preprint arXiv:1412.5567}.

\bibitem[{Miner et~al.(2020)Miner, Haque, Fries, Fleming, Wilfley, Terence~Wilson, Milstein, Jurafsky, Arnow, Stewart~Agras et~al.}]{miner2020assessing}
Adam~S Miner, Albert Haque, Jason~A Fries, Scott~L Fleming, Denise~E Wilfley, G~Terence~Wilson, Arnold Milstein, Dan Jurafsky, Bruce~A Arnow, W~Stewart~Agras, et~al. 2020.
\newblock Assessing the accuracy of automatic speech recognition for psychotherapy.
\newblock \emph{NPJ digital medicine}, 3(1):82.

\bibitem[{Radford et~al.(2023)Radford, Kim, Xu, Brockman, McLeavey, and Sutskever}]{radford2023robust}
Alec Radford, Jong~Wook Kim, Tao Xu, Greg Brockman, Christine McLeavey, and Ilya Sutskever. 2023.
\newblock Robust speech recognition via large-scale weak supervision.
\newblock In \emph{International conference on machine learning}, pages 28492--28518. PMLR.

\bibitem[{Vaswani et~al.(2017)Vaswani, Shazeer, Parmar, Uszkoreit, Jones, Gomez, Kaiser, and Polosukhin}]{Vaswani2017AttentionIA}
Ashish Vaswani, Noam~M. Shazeer, Niki Parmar, Jakob Uszkoreit, Llion Jones, Aidan~N. Gomez, Lukasz Kaiser, and Illia Polosukhin. 2017.
\newblock \href {https://api.semanticscholar.org/CorpusID:13756489} {Attention is all you need}.
\newblock In \emph{Neural Information Processing Systems}.

\bibitem[{Zhang et~al.(2023)Zhang, Han, Qin, Wang, Bapna, Chen, Chen, Li, Axelrod, Wang et~al.}]{zhang2023google}
Yu~Zhang, Wei Han, James Qin, Yongqiang Wang, Ankur Bapna, Zhehuai Chen, Nanxin Chen, Bo~Li, Vera Axelrod, Gary Wang, et~al. 2023.
\newblock Google usm: Scaling automatic speech recognition beyond 100 languages.
\newblock \emph{arXiv preprint arXiv:2303.01037}.

\end{thebibliography}




\end{document}